\documentclass{article}
\usepackage{tech2025}

\usepackage{microtype}
\usepackage{graphicx}
\usepackage{booktabs} 

\usepackage{hyperref}

\usepackage[utf8]{inputenc} 
\usepackage[T1]{fontenc}    
\usepackage{url}            
\usepackage{amsfonts}       
\usepackage{nicefrac}       
\usepackage[table]{xcolor}         
\usepackage{colortbl}
\usepackage{amsmath}
\usepackage{amssymb}
\usepackage{mathtools}
\usepackage{amsthm}
\usepackage{mathrsfs}
\usepackage{xspace}
\usepackage{academicons}

\usepackage{booktabs}
\usepackage{adjustbox}
\usepackage{algorithm}
\usepackage{algorithmicx}
\usepackage{amsmath,amsfonts}
\usepackage{bbm}
\usepackage{subcaption}  

\usepackage{pifont}

\usepackage{colortbl}
\usepackage{multirow}

\usepackage[capitalize,noabbrev]{cleveref}

\theoremstyle{plain}

\theoremstyle{definition}

\theoremstyle{remark}

\definecolor{bigaired}{RGB}{156, 0, 0}
\definecolor{uclablue}{RGB}{39, 116, 174}
\definecolor{thupurple}{RGB}{102, 8, 116}
\definecolor{pkured}{RGB}{139, 0, 18}
\definecolor{panton}{RGB}{217, 51, 121}

\definecolor{darkred}{RGB}{200, 0, 0}
\definecolor{darkblue}{RGB}{0, 0, 200}
\definecolor{blue}{RGB}{0, 0, 250}

\definecolor{light}{RGB}{225, 250, 250}
\definecolor{lightgray}{RGB}{0.9, 0.9, 0.9}
\definecolor{lightred}{RGB}{250, 200, 200}
\definecolor{lightblue}{RGB}{210, 220, 250}
\definecolor{lightpurple}{RGB}{218,210,255}

\definecolor{doderblue}{RGB}{30, 144, 255}
\definecolor{select}{RGB}{222, 235, 247}
\definecolor{unselect}{RGB}{247, 207, 206}

\definecolor{myLinkColor}{HTML}{7D5BA6}     
\definecolor{myCiteColor}{HTML}{9A4D92}     
\definecolor{myURLColor}{HTML}{5B7DB1}      
\hypersetup{
  colorlinks=true,
  linkcolor=myLinkColor,
  citecolor=myCiteColor,
  urlcolor=myURLColor
}

\usepackage[textsize=tiny]{todonotes}

\usepackage[most]{tcolorbox}

\usepackage{listings}

\usepackage{fontawesome5}  
\usepackage{listings}      
\usepackage[misc]{ifsym}
\usepackage{wrapfig}    
\usepackage[T1]{fontenc}

\usepackage{tcolorbox}
\usepackage{relsize}

\usepackage{adjustbox}
\usepackage{fancyhdr}
\usepackage{lipsum}
\usepackage{newtxtext}
\usepackage{wrapfig}
\usepackage{enumitem}
\definecolor{azblue}{RGB}{27,117,187}      

\usepackage[noend]{algpseudocode}   

\definecolor{bestcol}{RGB}{  0,102,204} 
\definecolor{goodcol}{RGB}{ 34,139, 34} 
\definecolor{deltaBg}{RGB}{220,230,255} 

\usepackage{lmodern}  

\usepackage{tikz}
\usetikzlibrary{tikzmark,decorations.pathreplacing,calc}
\usepackage{stackengine}
\stackMath
\definecolor{lightgreen}{RGB}{0,150,0}  

\usepackage{titletoc}

\newtheoremstyle{rqstyle}%
  {\topsep}            
  {\topsep}            
  {}                   
  {}                   
  {\bfseries}    
  {:}                  
  {.5em}               
  {}                   

\theoremstyle{rqstyle}

\crefname{researchquestion}{Research Question}{Research Questions}

\definecolor{propose}{HTML}{EF8E8D}
\definecolor{solve}{HTML}{5755A3}

\definecolor{humanred}{RGB}{180, 50, 50}
\definecolor{envgreen}{RGB}{50, 140, 80}

\definecolor{paleviolet}{HTML}{E1EEFC}
\definecolor{lightgrey}{RGB}{247, 247, 247}
\newenvironment{leapabstract}{
  \begin{tcolorbox}[
    colback=lightgrey,
    colframe=white,
    boxrule=0pt,
    arc=10pt,
    left=16pt,
    right=16pt,
    top=12pt,
    bottom=12pt,
    width=\textwidth,
    enlarge left by=0mm,
    before skip=10pt,
    after skip=10pt
  ]
  \normalsize
}{
  \end{tcolorbox}
}

\makeatletter
\DeclareRobustCommand\onedot{\futurelet\@let@token\@onedot}
\def\@onedot{\ifx\@let@token.\else.\null\fi\xspace}

\makeatother
\usepackage{makecell}   

\definecolor{lightgray}{rgb}{0.9,0.9,0.9}
\crefname{section}{Sec.}{Secs.}
\Crefname{section}{Sec.}{Secs.}
\crefname{subsection}{Sec.}{Secs.}
\Crefname{subsection}{Sec.}{Secs.}
\crefname{table}{Tab.}{Tabs.}
\Crefname{table}{Tab.}{Tabs.}
\crefname{figure}{Fig.}{Figs.}
\Crefname{figure}{Fig.}{Figs.}

\begin{document}

\makeatletter
\def\icmldate#1{\gdef\@icmldate{#1}}
\icmldate{\today}
\makeatother

\makeatletter
\fancypagestyle{fancytitlepage}{
  \fancyhead{}
  \lhead{\includegraphics[height=1.5cm]{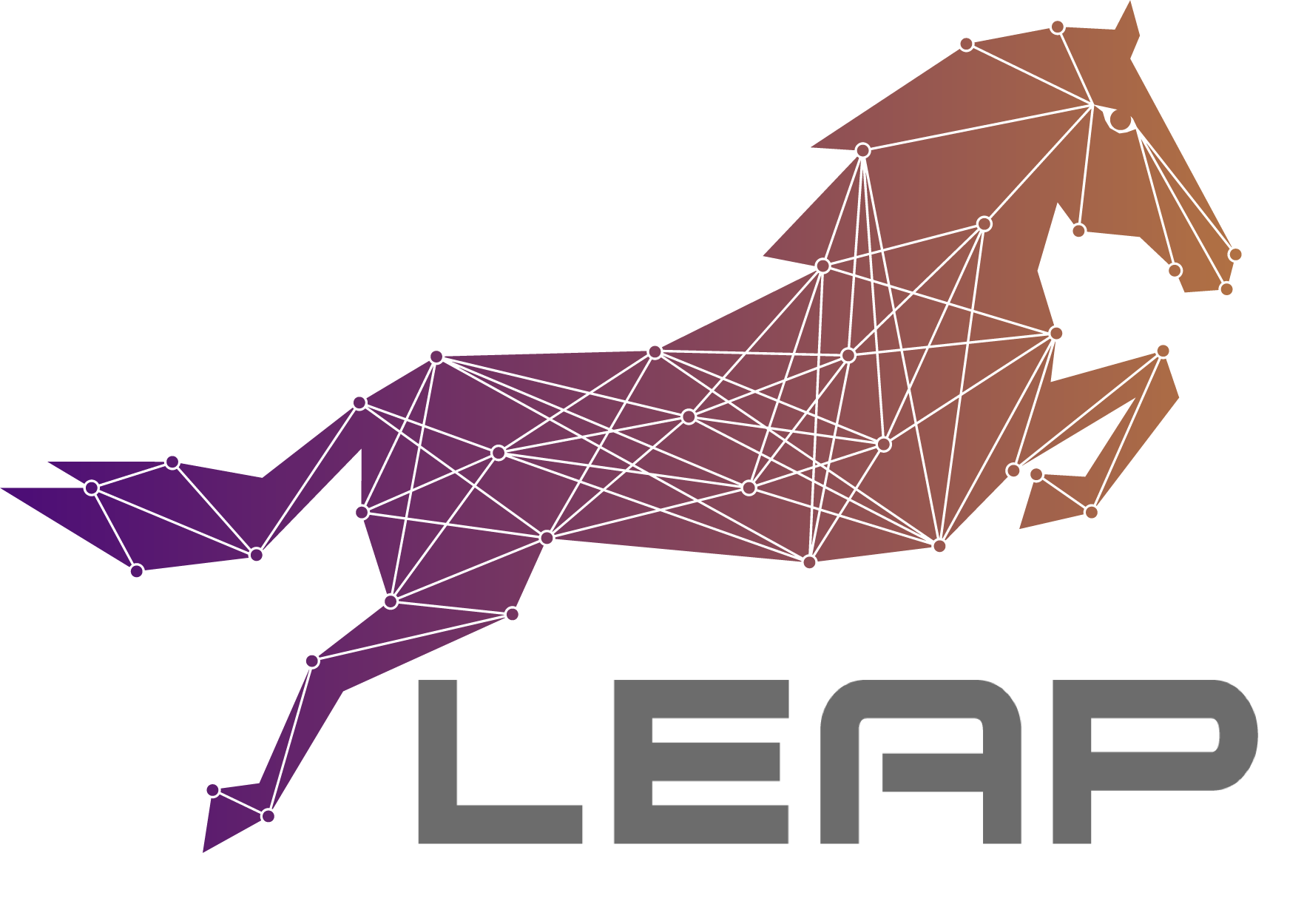}\hspace{5mm}}
  \rhead{\it \@icmldate}
  \cfoot{}
}
\makeatother

\icmltitle{What Makes World Action Models Generalize? An Empirical Study of Test-Time Future Modeling}

\begin{icmlauthorlist}
Renping Zhou$^{1,\ast}$, Zanlin Ni$^{1,\ast}$, Zihao Fan$^{2}$, Guohao Fu$^{3}$, Zeyu Liu$^{1}$, Hao Shi$^{1}$, Jie Zhang$^{1}$, Chi Bene Chen$^{1}$, Yang Yue$^{1}$, Xueyang Fu$^{2}$, Gao Huang$^{1,\textrm{\Letter}}$
\end{icmlauthorlist}

$^{1}$Leap Lab, Tsinghua University \ $^{2}$University of Science and Technology of China \ $^{3}$Beijing Institute of Technology
\icmlcorrespondingauthor{Renping Zhou}{tehaji007@gmail.com}

\vskip 0.1cm

\printNotice{}

\begin{leapabstract}
World action models (WAMs) predict the future alongside actions during \emph{training}. 
Due to the heavy computation cost of video denoising, whether the future must still be generated during \emph{inference} is disputed: Explicit WAMs denoise it into clean frames along with every action chunk, whereas Latent WAMs discard it entirely for acceleration. 
We find that latent WAMs, despite matching explicit ones on in-distribution tasks, fail to retain the generalization benefits that originally motivated WAMs. To demonstrate this, we evaluate generalization along three axes: \emph{environmental perturbation}, \emph{data efficiency}, and \emph{task generalization}. Controlled comparisons with a matched backbone, training data, and budget reveal consistent degradation across all three axes when the action expert no longer conditions on future representations.
Further analysis shows that the gap arises almost entirely from the first denoising step: the benefit comes from \emph{preparing} the future, not \emph{generating} it. 
We therefore propose \textbf{Simple-WAM}, which simplifies future modeling into a single forward pass of fully noised video tokens and adapts the training-time noise schedule to this inference behavior. 
Across simulation and real-world tasks, Simple-WAM achieves the best of both worlds, leading explicit WAMs in generalization performance with efficiency comparable to Latent WAMs. Project Page: \href{https://zrporz.github.io/Simple-WAM-Web/}{\textcolor{panton}{\texttt{https://zrporz.github.io/Simple-WAM-Web}}}

\end{leapabstract}


\section{Introduction}
\label{sec:intro}

Building generalizable robot policies is a long-standing goal of embodied AI.
World Action Models (WAMs) pursue it by jointly generating future dynamics and actions conditioned on observations and language instructions, which supplies supervision in observation space beyond sparse action labels~\cite{ye2026world,li2026causal,Bi_2026_CVPR,kim2026cosmos}.
Initialized from video generation models trained on web-scale video data, WAMs inherit rich spatiotemporal priors and shift action learning from dense state-action imitation toward inverse dynamics that aligns motor commands with predicted visual futures, and it is this capacity to model the future that is credited with their stronger robustness and generalization~\cite{cheang2024gr2generativevideolanguageactionmodel,pai2025mimic,hu2024video}.
This sets them apart from Vision-Language-Action (VLA) models~\cite{brohan2023rt,black2024pi,intelligence2025pi,liu2025rdt,kim2024openvla,ICLR2026_1fa3d6cc}, whose backbones are pretrained predominantly on static image-text pairs and optimized for understanding or reasoning rather than for generation, and which therefore carry little of the dynamic understanding of temporal evolution that manipulation demands~\cite{zhou2026exploring,guruprasad2025benchmarking,ye2026gigaworld}.

While the value of future modeling to WAMs during training is widely accepted, whether the future must also be modeled at inference, and through what mechanism it would then act on the action, is still an underexplored question.
Early WAMs~\cite{hu2024video,cheang2024gr2generativevideolanguageactionmodel,ye2026world,kim2026cosmos,pai2025mimic} adopt the \emph{explicit paradigm}, in which the future is denoised into clean frames alongside every action chunk and the action is conditioned on them, and report that the success rate tracks the quality of that generated future, which established it as an indispensable component of inference~\cite{ye2026world,li2026causal}.
Because video tokens far outnumber action tokens, however, that denoising dominates the per-chunk cost and hinders high-frequency closed-loop control. 
More recent works~\cite{yuan2026fast,ye2026gigaworld} focused on efficiency argue instead that future modeling contributes to the policy mainly as a training objective rather than as a test-time requirement, and propose a \emph{latent paradigm} that retains video supervision during training but skips the heavy video generation at inference, reporting little difference from explicit WAMs at a fraction of the cost.
Motivated by these discussions, we raise the following question: is inference-time video generation necessary after all?

\begin{figure}[t]
  \centering
  \includegraphics[width=\linewidth]{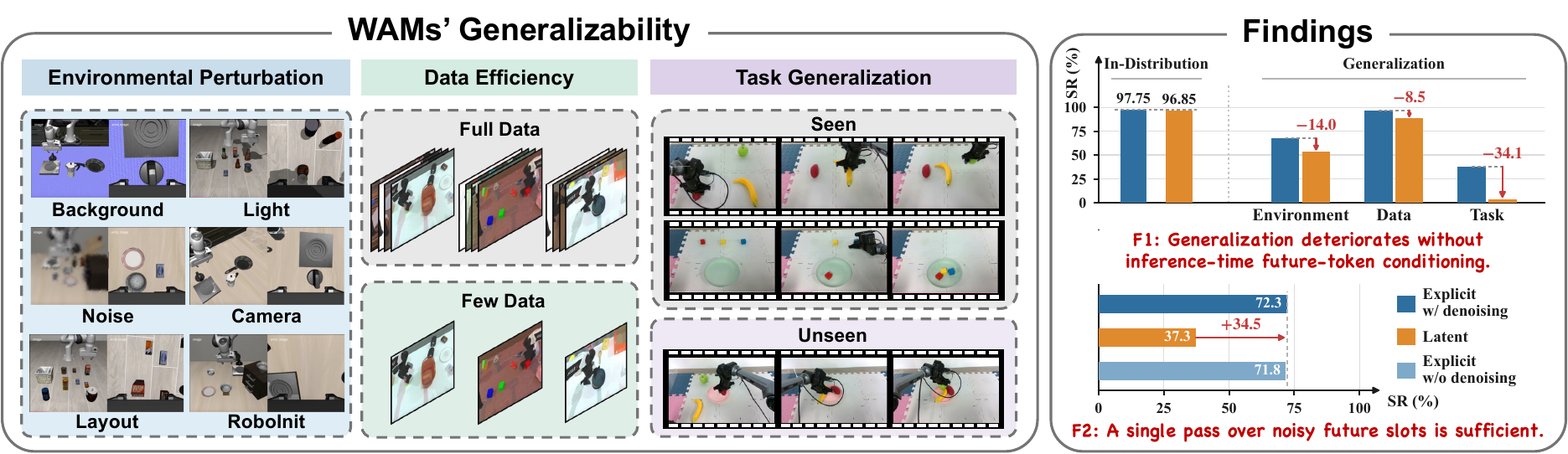}
  \caption{\textbf{An empirical study of test-time future modeling for WAM
  generalization.}
  \textbf{Left:} the three axes of the protocol, environmental perturbation,
  data efficiency, and task generalization.
  \textbf{Right:} the explicit and latent paradigms are within $0.9$ points on in-distribution tasks yet separate on all three axes (F1), and a single forward pass
  over a fully noised future recovers almost all of that separation (F2).}
  \label{fig:teaser}
  \vspace{-10pt}
\end{figure}

We answer this through a controlled comparison of the two paradigms, and find that neither claim is wrong on the evidence that supports it. Latent WAMs match explicit ones on in-distribution tasks at much lower inference cost, but this parity does not imply comparable generalization that originally motivated WAMs. To examine this, we break down WAMs' generalizability into three complementary axes and propose an evaluation framework organized around them: \emph{robustness to environmental perturbations}, \emph{data efficiency with fewer demonstrations}, and \emph{generalization to tasks unseen in action training}.


Under this framework, we compare matched configurations that vary whether and in what form the action expert accesses future-token representations. Our study yields two findings, demonstrated in~\cref{fig:teaser}. \emph{(i) Generalization deteriorates without inference-time future modeling.} Under our matched setting, latent WAMs match explicit WAMs in distribution, yet fall behind across all three generalization axes. \emph{(ii) Fully noised future is sufficient for generalization.} A single forward pass over fully noised future video tokens produces representations that recover most of the gap, without iterative denoising into clean future frames. This indicates that most of the generalization benefit can be retained through inference-time future conditioning without generating a clean future.


Guided by these findings, we propose \textbf{Simple-WAM}, which restores inference-time future conditioning to latent WAMs at negligible additional cost.
Simple-WAM retains the single-pass prefill and inference cost of the latent paradigm, but keeps the future in the policy's context as future video tokens left at noise.
At inference, the video expert runs once instead of over the whole schedule, and training shifts its flow-time sampling toward that same point. As a result, neither stage pays for the denoising Finding~2 shows to be unnecessary.
Simple-WAM leads the explicit paradigm on seven of the eight measurements of \cref{tab:main} and the latent one on all eight, reaching $79.5$ under environmental perturbation against $67.7$ and $53.8$, while running at $3.8\times$ the speed of the explicit paradigm and within $1.2\times$ the latency of the latent one. 
Therefore, the choice between the two paradigms is not the trade-off between generalization and efficiency it is usually taken to be. 


\section{Related Works}
\label{sec:related}

\subsection{World Action Models and Inference Efficiency}

Using generated video as an intermediate for control predates the current formulation: early predictive policies imagine a future from the current observation and recover the action from it by inverse dynamics~\cite{du2023learning,zhou2024robodreamer,hu2024video,shi2026memoryvla++}, and large-scale video pretraining was shown to transfer to manipulation ahead of any action label~\cite{wu2023unleashing,cheang2024gr2generativevideolanguageactionmodel}.
World action models tighten this coupling into a single network that predicts future frames and actions jointly~\cite{ye2026world,li2026causal,Bi_2026_CVPR,kim2026cosmos,pai2025mimic}.
They attribute the advantage to a training signal defined over the observation rather than the action alone, and to a video backbone carrying a prior on how a scene evolves.
These works demonstrate gains in sample efficiency, robustness to scene variation, and transfer to tasks and embodiments beyond the action data~\cite{ye2026world,pai2025mimic,li2026causal}.

That prediction is paid for at inference, where the video branch is denoised once per action chunk over far more tokens than the action branch.
A number of recent works therefore examine how an efficient WAM should be built, along the model architecture~\cite{li2026light,zhao2026faster}, the inference-time formulation~\cite{yuan2026fast,zhang2026keep,ye2026gigaworld}, and the training recipe~\cite{li2026efficient,luo2026being,zhang2026imagewam}.
Among them, the latent paradigm makes the strongest claim: that future prediction contributes as a training objective rather than as a test-time requirement, so its video branch is trained as usual but never asked to produce a future at deployment~\cite{yuan2026fast,ye2026gigaworld}.
Success rates close to the explicit paradigm's are reported at a fraction of its latency, so whether the future must be generated at inference is still contested.
That evidence is, however, collected in distribution.
Concurrent work such as Faster-WAM~\cite{zhao2026faster} highlights the importance of inference-time future conditioning for robustness under environmental perturbations and develops an efficient implementation. Our work differs primarily in establishing a broader evaluation framework spanning environmental perturbation, data efficiency, and task generalization, and conducting controlled experiments within this framework to examine how future conditioning and iterative video denoising affect generalization.


\subsection{Generalization of Embodied Policies}

Generalization is a key ability for embodied agents, and VLA and WAM policies alike are evaluated on it not as a single quantity but along dimensions that differ in what changes at deployment~\cite{liu2023libero,chen2025robotwin,guruprasad2025benchmarking}.
Closest to the training distribution is variation that leaves the task unchanged, which LIBERO-Plus decomposes into seven factors covering the camera, the scene, the language, and the robot's initial state~\cite{fei2026liberoplus}.
Holding the task fixed and reducing its supervision gives data efficiency, measured by success under fewer demonstrations per task, where video-pretrained policies report their clearest gains~\cite{pai2025mimic,li2026causal,ye2026world}.
Furthest is the case in which the task itself changes, studied as transfer to tasks withheld from training~\cite{zhou2026exploring,guruprasad2025benchmarking} or as the acquisition of a skill from action-free video of it~\cite{ye2026world,li2026causal}.
These properties are claimed or observed across prior WAM work, but each is demonstrated by one system under its own backbone, data, and training budget, and the dimensions are seldom placed side by side.
We consolidate them into three axes and compare the paradigms under one matched setting that isolates inference-time future modeling, so as to ask what makes WAMs generalize and how that generalization can be retained at the smallest inference cost.

\section{Preliminaries}
\label{sec:prelim}

%

\subsection{World Action Models}

We consider language-conditioned visuomotor control from demonstrations.
At control step $t$, the policy receives one or more image observations $o_t$, a language instruction $\ell$, and a proprioceptive state $s_t$, and predicts an action chunk $A_t=a_{t:t+H-1}$ of horizon $H$.
The language instruction $\ell$ and proprioceptive state $s_t$ condition every stage of the model. For notational simplicity, we omit them below.

A world action model couples a video expert with parameters $\theta$, initialized from a pretrained video generator, and an action expert with parameters $\psi$~\cite{ye2026world,li2026causal,Bi_2026_CVPR,kim2026cosmos}.
The video expert reads a single sequence of latents spanning the current and future frames.
Let $\mathcal{E}$ be the visual encoder of the backbone and $y_{t}=\mathcal{E}(o_t)$ the latent of the current observation, which is given and therefore clean, while the latents $y_{t+1:t+T}$ of the $T$ future frames are unknown at inference and carry a flow time $\tau\in[0,1]$, with $\tau=0$ at clean data and $\tau=1$ at Gaussian noise.
Writing $y^{\tau}_{t+1:t+T}=(1-\tau)\,y_{t+1:t+T}+\tau\epsilon$ for the interpolant, the video expert operates on
\begin{equation}
  Y_t^{\tau}
  =
  \bigl[\,y_{t}\;;\;y^{\tau}_{t+1:t+T}\,\bigr],
  \label{eq:video_seq}
\end{equation}
and regresses the velocity field at the future video tokens,
\begin{equation}
  \mathcal{L}_{\mathrm{vid}}
  =
  \mathbb{E}_{\tau,\epsilon}
  \left\lVert
    u_{\theta}\!\left(Y_t^{\tau},\tau\right)-\left(\epsilon-y_{t+1:t+T}\right)
  \right\rVert^{2}.
  \label{eq:vid_loss}
\end{equation}
The action expert regresses its own velocity field $v_{\psi}$ over an independent flow time $\sigma$ and interpolant $A_t^{\sigma}$, and training minimizes $\mathcal{L}=\mathcal{L}_{\mathrm{act}}+\lambda\,\mathcal{L}_{\mathrm{vid}}$.
Both flow times follow a schedule $\mathcal{S}$ that serves training and inference alike: the expectation over $\tau$ in Eq.~\eqref{eq:vid_loss} draws $\tau=\mathcal{S}(u)$ with $u\sim\mathcal{U}[0,1]$, and the steps $\tau_k$ of \cref{sec:prelim_conditioning} are the same schedule discretized.
Because $\tau$ and $\sigma$ are sampled independently, the action expert is trained against future latents at every noise level, including $\tau$ near $1$.

\subsection{Inference-Time Future Conditioning}
\label{sec:prelim_conditioning}

At inference the action chunk is produced by integrating the action velocity field from $\sigma_1=1$ to $\sigma_{K+1}=0$ over $K$ steps, and the executed chunk is the resulting $A_t^{0}$.
Let $\Phi_{\theta}$ denote the map from the video expert to the representation read by the action expert.
At step $k$ the action latent is updated by
\begin{equation}
  A_t^{\sigma_{k+1}}
  =
  A_t^{\sigma_{k}}
  +
  \left(\sigma_{k+1}-\sigma_{k}\right)
  v_{\psi}\!\left(A_t^{\sigma_{k}},\,c_{t,k},\,\sigma_{k}\right),
  \label{eq:action_step}
\end{equation}
where the future conditioning feature supplied at that step is
\begin{equation}
  c_{t,k}
  =
  \Phi_{\theta}\!\left(Y_t^{\tau_k}\right).
  \label{eq:cond}
\end{equation}
Both paradigms are instances of Eq.~\eqref{eq:action_step}, and two quantities in Eq.~\eqref{eq:cond} distinguish them: whether future video tokens are present in the video sequence, and what schedule the flow time $\tau_k$ of those positions follows.

\noindent\textbf{Explicit WAMs}
keep the future video tokens and drive their flow time toward clean data,
\begin{equation}
  c_{t,k}=\Phi_{\theta}\!\left(Y_t^{\tau_k}\right),
  \qquad
  \tau_1=1\;\longrightarrow\;\tau_{K+1}=0 .
  \label{eq:c_explicit}
\end{equation}
The action expert is therefore conditioned on a progressively sharper future rather than on a single fully denoised one: at every step it reads the future at that step's noise level.
Jointly denoising models step the video and action experts along this shared schedule~\cite{Bi_2026_CVPR,li2026causal,ye2026world,kim2026cosmos}.
Generate-then-act models are the special case in which the video is denoised to $\tau=0$ first and the action expert reads $\Phi_{\theta}(Y_t^{0})$ at every step. 

\noindent\textbf{Latent WAMs}
drop the future video tokens from the sequence at inference while retaining future prediction during training~\cite{yuan2026fast,ye2026gigaworld},
\begin{equation}
  c_{t,k}=\Phi_{\theta}\!\left(\bigl[\,y_{t}\,\bigr]\right)
  \quad\text{for all }k .
  \label{eq:c_latent}
\end{equation}
The video expert still makes one forward pass, on the current-frame latent alone, but no flow time is defined and the feature read by the action expert is constant across the $K$ steps.

\noindent\textbf{Inference cost.}
Let $C_{\mathrm{vid}}$ be the cost of one video-expert forward pass carrying future video tokens, $C_{\mathrm{cur}}$ the cost of one pass on $[\,y_t\,]$ alone, and $C_{\mathrm{act}}$ the cost of one action step.
Reaching $\tau_{K+1}=0$ requires denoising the video over the whole schedule, so the explicit paradigm costs $K\,C_{\mathrm{vid}}+K\,C_{\mathrm{act}}$, whereas the latent paradigm costs $C_{\mathrm{cur}}+K\,C_{\mathrm{act}}$.
Two factors make $C_{\mathrm{vid}}$ large relative to $C_{\mathrm{act}}$.
The video expert carries the future video tokens and therefore a far longer sequence than the action expert, $\lvert y_{t+1:t+T}\rvert\gg\lvert A_t\rvert$, and it is also the larger of the two, since it is initialized from a pretrained video generator while the action expert is comparatively small. 
The video term therefore dominates in the explicit case and is the source of the latency gap reported in \cref{sec:method}.
The cost is set by how far $\tau_k$ is driven toward $0$, not by how many times the action expert reads the future: a schedule held near $\tau=1$ requires no denoising at all.

Prior work has compared only these two settings, which differ in both factors at once.
Our study holds every other component of Eq.~\eqref{eq:action_step} fixed and varies them separately.

\section{A Controlled Study of Inference-Time Future Conditioning}
\label{sec:study}


\subsection{Evaluation Protocol for Generalizability}
\label{sec:protocol}

The generalization benefits of WAMs have been widely discussed~\cite{ye2026world,pai2025mimic,li2026causal}.
However, a systematic evaluation of this capability, and a controlled comparison across paradigms and components, are still lacking.
We therefore propose a protocol including three axes, as shown in~\cref{fig:teaser}, to systematically evaluate the generalization capabilities of WAMs.

\noindent\textbf{Environmental perturbation.}
Trained to predict how a scene evolves and initialized from video models carrying rich spatiotemporal priors, WAMs are expected to remain robust under conditions not observed during training~\cite{ye2026world}.
We evaluate the in-distribution checkpoint without retraining, under the seven perturbation factors of LIBERO-Plus~\cite{fei2026liberoplus}: robot initial states, camera viewpoints, language instructions, sensor noise, backgrounds, object layouts, and lighting.

\noindent\textbf{Data efficiency.}
WAMs are reported to retain their competence when the demonstrations available
per task are reduced~\cite{ye2026world,pai2025mimic,kim2026cosmos}.
Such data efficiency is attributed to video pretraining, which already supplies
the physical dynamics of how objects move and interact.
LIBERO provides 40 to 50 demonstrations per task; we retrain each configuration
from the same initialization with the per-task count reduced to 10, and evaluate
in distribution to isolate the impact of demonstration count.

\noindent\textbf{Task generalization.}
WAMs trained across diverse tasks are reported to acquire skills for unseen tasks without any additional data or merely by seeing the operation video~\cite{ye2026world,li2026causal}.
We evaluate this ability under two settings: \emph{without video}, where no data of the held-out tasks is available at any stage, and \emph{with video}, where video of those tasks is available but carries no action labels. We use it to train the video branch only.
Both settings use four-fold cross-validation over the four LIBERO suites, spatial, object, goal, and long~\cite{liu2023libero}: each fold withholds one suite and trains on the remaining three, and we report the average success rate.

To compare the paradigms rather than their implementations, every configuration
in this section follows the architecture and training hyperparameters of
Fast-WAM~\cite{yuan2026fast}: a pretrained Wan2.2-5B video DiT~\cite{wan2025}
as the video backbone, reusing its text encoder and video VAE, and a
$1$B action expert initialized from the interpolation of the video DiT.
Following Fast-WAM, we use the same flow-time sampling strategy during training and the corresponding discretized schedule during inference.
The explicit and latent paradigms differ only in a structured attention mask,
which decides whether the action tokens may attend to the future video tokens.
On each of the three axes the training budget is the same as in distribution.


\subsection{Finding 1:  Generalization Deteriorates without Inference-time Future Modeling}
\label{sec:finding_1}

\definecolor{dropgrey}{gray}{0.55}
\newcommand{\idg}[1]{\textcolor{dropgrey}{#1}}
\begin{wraptable}{r}{0.47\textwidth}
  \centering
  \vspace{-\intextsep}
  \footnotesize
  \caption{\textbf{Generalization Deteriorates without Inference-time Future Modeling.}
  Success rate (\%) under the matched setting of \cref{sec:protocol}.
  w/ vid: action-free video of the held-out tasks with video available.
  $\Delta$ is Explicit minus Latent; the in-distribution column is
  \idg{greyed}, and the generalization deltas are \textbf{bold}.}
  \label{tab:finding1}
  \setlength{\tabcolsep}{2.6pt}
  \begin{tabular}{lccccc}
    \toprule
    & \multirow{2}{*}{\idg{In-dist.}} & \multicolumn{4}{c}{Generalization} \\
    \cmidrule(lr){3-6}
    & & Perturb. & Data Eff. & \multicolumn{2}{c}{Task Gen.} \\
    \cmidrule(lr){3-3}\cmidrule(lr){4-4}\cmidrule(lr){5-6}
    Paradigm & \idg{LIBERO} & LIBERO-Plus & 10-shot & w/o vid & w/ vid \\
    \midrule
    Explicit & \idg{97.75} & 67.72 & 96.95 & 6.25 & 69.90 \\
    Latent   & \idg{96.85} & 53.75 & 88.50 & 2.10 & 5.90 \\
    \midrule
    $\Delta$ & \idg{$0.90$}
    & $\mathbf{13.97}$ & $\mathbf{8.45}$ & $\mathbf{4.15}$ & $\mathbf{64.00}$ \\
    \bottomrule
    \end{tabular}
\end{wraptable}

We first ask whether the future can be removed at inference without cost.
\cref{tab:finding1} compares the explicit and latent paradigms.
In distribution the two are comparable, at $97.75$ against $96.85$, in line with what latent WAM works report~\cite{ye2026gigaworld,yuan2026fast}.
However, a clear gap emerges under the three axes of our protocol: the explicit paradigm leads by $13.97$ points under environmental perturbation and by $8.45$ under data efficiency.
Task generalization shows the largest separation.
With no data of the held-out tasks both paradigms stay near the floor, the explicit one slightly ahead.
Acquiring a skill from action-free video is one of the properties claimed for WAMs~\cite{ye2026world,li2026causal}, and only the explicit paradigm realizes it: such video is worth $63.65$ points to it and $3.80$ to the latent paradigm, which reaches $5.90$ against $69.90$.

The gap shows that an in-distribution match is not sufficient for comparing the two paradigms. The latent paradigm removes future modeling at inference to reduce cost, and loses generalization as a result. Future modeling is therefore a requirement at inference, not only a training objective.

\subsection{Finding 2: Fully Noised Future is Sufficient for Generalization}
\label{sec:findings_probe}


Finding~1 establishes that the future must be modeled at inference for a WAM to generalize, and leaves open how much of it is required.
A natural question is whether a future denoised more clearly yields a stronger policy, and what kind of future is sufficient for that generalization.
The same question matters for efficiency, since denoising the future dominates the per-chunk cost~(\cref{sec:prelim_conditioning}).

\begin{wrapfigure}{r}{0.47\textwidth}
  \centering
  \vspace{-\intextsep}
  \includegraphics[width=\linewidth]{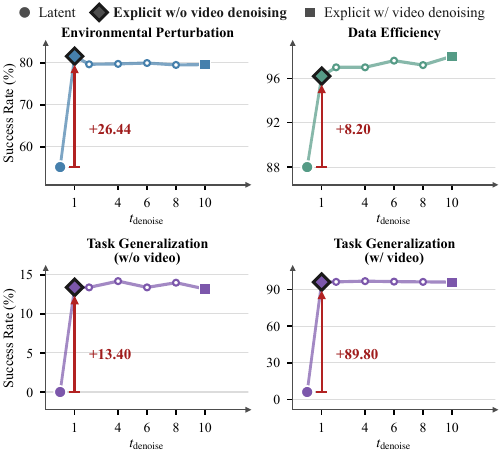}
  \caption{\textbf{Fully Noised Future Is Sufficient for Generalization.}
  Success rate (\%) on LIBERO-Spatial for a model trained under the explicit
  paradigm. The video expert is run for $t_{\mathrm{denoise}}$ of the $K$ denoising steps;
  no retraining is performed.}
  \label{fig:finding2}
\end{wrapfigure}

To answer this question, we conduct an experiment that changes how far the future is denoised.
We take a model trained under the explicit paradigm and run its video expert for only the first $t_{\mathrm{denoise}}$ of the $K$ steps of the schedule, holding the feature it supplies fixed for the rest of the action chunk, so that $c_{t,k}=c_{t,t_{\mathrm{denoise}}}$ for every $k>t_{\mathrm{denoise}}$.
$t_{\mathrm{denoise}}$ counts how many times the video expert is run before the action expert stops seeing the future change.
It is set at inference alone: nothing is retrained, the weights and the action denoising are those of the explicit paradigm, and $t_{\mathrm{denoise}}$ is the only quantity that varies.
At $t_{\mathrm{denoise}}=K$ this recovers Eq.~\eqref{eq:c_explicit}, the explicit paradigm itself.

The sweep is reported in \cref{fig:finding2}, and almost all of the gain comes from the first pass.
At $t_{\mathrm{denoise}}=1$ the video expert is run once, at $\tau_1=1$, where $y^{\tau}_{t+1:t+T}=\epsilon\sim\mathcal{N}(0,I)$: the future video tokens of Eq.~\eqref{eq:video_seq} are pure Gaussian noise, so nothing about the future has been explicitly denoised when the action expert conditions on it.
Even so, this single pass outperforms the latent paradigm by $26.44$, $8.2$, $13.4$ and $89.8$ points across the four settings.
The nine forward passes that follow, which carry the entire cost of denoising the video, reach at best $-1.62$, $+1.8$, $+0.8$ and $+0.8$ against it.

The gain therefore does not come from the future the video expert generates, but from the single pass that forms the intermediate representation the action expert reads.
This implies a misalignment between the visual fidelity the video model is trained for and the feature the action expert requires: a fully noised future is already sufficient.


%

\section{Method}
\label{sec:method}

\begin{figure}[t]
  \centering
  \includegraphics[width=0.95\textwidth]{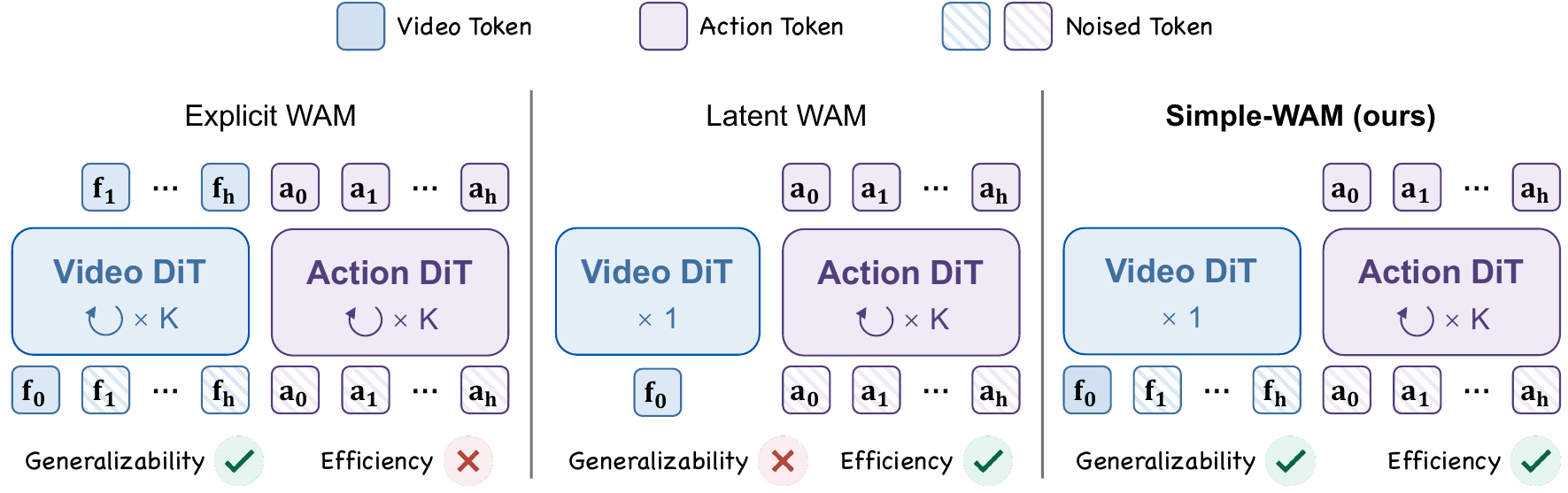}
  \caption{\textbf{The three conditioning schemes.}
  The explicit paradigm denoises the future video tokens across the schedule;
  the latent paradigm drops them and keeps only the single forward pass.
  We propose Simple-WAM, which keeps both the future video tokens and the single forward pass: future video tokens are input as pure Gaussian noise at
  $\tau=1$ and are never denoised.}
  \label{fig:model}
\end{figure}

Guided by two findings in \cref{sec:study}, in this section, we propose \textbf{Simple-WAM}, which makes two changes to the explicit paradigm, and keeps only the part of the video expert that the two findings call for.
Since each denoising step runs the video expert once, dropping the steps Finding~2 shows to be unnecessary removes most of the inference cost.
Under Eq.~\eqref{eq:cond}, Simple-WAM balances the two paradigms: the future
video tokens are conditioned on, as in the explicit one, and the video expert makes a single forward pass, as in the latent one (\cref{fig:model}).
Neither change adds a module or a loss term, and together they approach the inference cost of a latent WAM with the generalization of an explicit one.

\subsection{Inference: One Pass over a Fully Noised Future}
\label{sec:method_infer}
During inference,  Simple-WAM gives up the iterative denoising of the future video tokens from the explicit paradigm. The video expert makes one forward pass over the sequence with its future video
tokens left at Gaussian noise, and the feature it produces conditions all $K$
action steps,
\begin{equation}
  c_{t,k}=\Phi_{\theta}\!\left(Y_t^{\tau=1}\right)
  \quad\text{for all }k .
  \label{eq:c_lf}
\end{equation}
The action denoising of Eq.~\eqref{eq:action_step} is unchanged.
Against the two paradigms of~\cref{sec:prelim_conditioning}, this keeps the
future video tokens the explicit one reads and pays only for the single forward
pass the latent one makes: $C_{\mathrm{vid}}+K\,C_{\mathrm{act}}$, against
$K\,C_{\mathrm{vid}}+K\,C_{\mathrm{act}}$ and
$C_{\mathrm{cur}}+K\,C_{\mathrm{act}}$.

\subsection{Training: A Mixed Schedule}
\label{sec:method_train}

\cref{sec:method_infer} rests the entire future feature on an input at $\tau=1$, while
the explicit paradigm trains its video expert by denoising across the whole
schedule. The share of the training budget spent at $\tau=1$ is therefore
negligible, while at inference it is the only flow time used. To strengthen the
ability
to read a feature out of pure noise, we raise that share with a single scalar
$p$ and draw the flow time of the future video tokens as
\begin{equation}
  \tau =
  \begin{cases}
    1, & \text{with probability } p,\\[2pt]
    \mathcal{S}(u),\quad u\sim\mathcal{U}[0,1], & \text{otherwise,}
  \end{cases}
  \label{eq:tau_mix}
\end{equation}
where $\mathcal{S}$ is the flow-time schedule of the explicit paradigm.
Training is otherwise unchanged: the two experts are trained jointly as
in~\cref{sec:prelim}, the action expert reads the future video tokens through
$\Phi_{\theta}$, and $\mathcal{L}_{\mathrm{vid}}$ is applied at the sampled
$\tau$. The remaining $(1-p)$ keeps the video expert supervised across the rest
of the schedule.
The mixture therefore sharpens the feature the video expert produces at the one
flow time inference reads, while $(1-p)$ leaves it under the flow-time sampling
it was originally trained with.
We set $p=0.5$ and ablate it in~\cref{sec:exp_ablation}. 

\section{Experiments}
\label{sec:exp}


\subsection{Implementation Details}
\label{sec:exp_impl}

Simple-WAM is built on the setting of~\cref{sec:protocol}. We use
pretrained Wan2.2-5B~\cite{wan2025} as the video backbone, set the action chunk
to $H=32$, and use 9 video frames per chunk. During training we set the video
loss weight to $\lambda=3.0$. And we use $K=10$ denoising steps
with a classifier-free guidance (CFG) scale set to $1.0$ at inference time. All remaining training
settings follow Fast-WAM~\cite{yuan2026fast}. All models are trained on NVIDIA
A800 GPUs, and latency numbers are measured on a single NVIDIA GeForce
RTX 5090 GPU under each configuration's own denoising settings. We provide more hyperparameter settings in~\cref{app:training_details}.

\newcommand{\ep}{\ding{51}}
\newcommand{\noep}{\ding{55}}
\definecolor{refgray}{gray}{0.58}
\newcommand{\gy}[1]{\textcolor{refgray}{#1}}
\begin{table}[t]
  \centering
  \footnotesize
  \caption{\textbf{Main results across the three generalization axes.}
  Success rate (\%) under the protocol of \cref{sec:protocol}, per factor for
  environmental perturbation and averaged over suites elsewhere. w/o vid.\ and
  w/ vid.\ denote task generalization without and with action-free video of the
  held-out tasks; such video  provides no training signal in a VLA, so ``*'' repeats the
  w/o vid.\ entry. ``Emb.\ P.T.'' denotes large-scale embodied pretraining; those
  rows are \textcolor{refgray}{grey}, and \textbf{bold} indicates the best among the
  rest. ``--'': not covered by the official release. Detailed results are in \cref{tab:per_suite}.}
  \label{tab:main}
  \setlength{\tabcolsep}{1.0pt}
  \begin{tabular}{lcccccccccccccccc}
    \toprule
    & & \multicolumn{8}{c}{Environmental perturbation}
    & \multicolumn{3}{c}{Data efficiency}
    & \multicolumn{4}{c}{Task generalization} \\
    \cmidrule(lr){3-10}\cmidrule(lr){11-13}\cmidrule(lr){14-17}
    & & \multicolumn{8}{c}{LIBERO-Plus}
    & \multicolumn{2}{c}{LIBERO}
    & RoboTwin
    & \multicolumn{2}{c}{LIBERO}
    & \multicolumn{2}{c}{RoboTwin} \\
    \cmidrule(lr){3-10}\cmidrule(lr){11-12}\cmidrule(lr){13-13}\cmidrule(lr){14-15}\cmidrule(lr){16-17}
    Method & \makecell{Emb.\\P.T.}
    & Init. & Cam. & Lang. & Noise & Bg. & Lay. & Light & Avg.
    & 5-shot & 10-shot & 10-shot
    & w/o vid & w/ vid & w/o vid & w/ vid \\
    \midrule
    \rowcolor{lightgray}\multicolumn{17}{l}{\textit{Vision-language-action models}} \\
    \hspace*{0.6em}\gy{$\pi_{0.5}$}~\citeyearpar{intelligence2025pi} & \ep
    & \gy{73.6} & \gy{78.4} & \gy{80.8} & \gy{89} & \gy{94.1} & \gy{84.5} & \gy{96.2} & \gy{84.4}
    & \gy{87.5} & \gy{91.5} & \gy{33.9}
    & \gy{9.4} & \gy{9.4*} & \gy{2.5} & \gy{2.5*} \\
    \hspace*{0.6em}VLA-Adapter~\citeyearpar{wang2025vlaadapter} & \noep
    & 37.4 & 36.4 & 73.8 & 57.2 & \textbf{76.6} & 70.2 & 71.0 & 59.0
    & 77.6 & 85.9 & --
    & 0.3 & 0.3* & -- & -- \\
    \addlinespace[2pt]
    \rowcolor{lightgray}\multicolumn{17}{l}{\textit{Explicit WAMs}} \\
    \hspace*{0.6em}\gy{Lingbot-VA}~\citeyearpar{li2026causal} & \ep
    & \gy{83.0} & \gy{86.4} & \gy{82.3} & \gy{53.1} & \gy{40.9} & \gy{64.4} & \gy{76.2} & \gy{70.5}
    & \gy{78.5} & \gy{87.6} & \gy{3.9}
    & \gy{15.2} & \gy{71.1} & \gy{0.0} & \gy{1.1} \\
    \hspace*{0.6em}FastWAM-Joint~\citeyearpar{yuan2026fast} & \noep
    & 64.9 & 36.4 & 94.8 & 54.8 & 54.7 & 78.6 & 94.9 & 67.7
    & 91.5 & 97.0 & 35.0
    & 6.3 & 69.9 & 5.4 & \textbf{47.3} \\
    \addlinespace[2pt]
    \rowcolor{lightgray}\multicolumn{17}{l}{\textit{Latent WAMs}} \\
    \hspace*{0.6em}FastWAM~\citeyearpar{yuan2026fast} & \noep
    & 49.5 & 21.0 & 73.3 & 46.7 & 52.0 & 63.7 & 77.5 & 53.8
    & 78.2 & 88.5 & 4.8
    & 2.1 & 5.9 & 3.5 & 4.8 \\
    \addlinespace[3.5pt]
    \textbf{Simple-WAM} & \noep
    & \textbf{82.1} & \textbf{55.6} & \textbf{96.2} & \textbf{74.2} & 72.9 & \textbf{83.4} & \textbf{95.8} & \textbf{79.5}
    & \textbf{92.4} & \textbf{97.2} & \textbf{37.2}
    & \textbf{10.1} & \textbf{73.6} & \textbf{6.2} & 44.5 \\
    \bottomrule
  \end{tabular}
\end{table}

\subsection{Experimental Setup}
\label{sec:exp_setup}

We evaluate Simple-WAM on LIBERO~\cite{liu2023libero}, RoboTwin 2.0~\cite{chen2025robotwin}, and LIBERO-Plus~\cite{fei2026liberoplus}, along the
three axes of~\cref{sec:protocol}.

\noindent\textbf{LIBERO.}
LIBERO comprises four suites covering spatial relations, object-centric skills,
goal-conditioned tasks, and long-horizon behaviors, each with 10 tasks and 500
expert demonstrations. We follow the standard protocol in distribution, and
form the other two axes as in~\cref{sec:protocol}: for data efficiency we
retrain with the demonstrations per task reduced to 10 and to 5, and for task
generalization we run four-fold cross-validation over the suites.

\noindent\textbf{RoboTwin 2.0.}
RoboTwin 2.0 is a bimanual manipulation benchmark with tasks that
require coordinated dual-arm control, reported under both clean and randomized
scenes. We repeat the same two axes here: for data efficiency we retrain with
10 demonstrations per task, and for task generalization we withhold 10 randomly
selected tasks.

\noindent\textbf{LIBERO-Plus.}
LIBERO-Plus extends the LIBERO tasks with seven kinds of variation, which form the environmental perturbation axis
of \cref{sec:protocol}. The overall score is computed by weighting the seven variation factors according to their respective task counts, following the official evaluation protocol.


\noindent\textbf{Real-World Evaluation.}
We conduct real-world experiments on an AgileX Aloha dual-arm platform.
We consider four tasks, \textit{Stack the Bowls}, \textit{Store the Blocks},
\textit{Sort in Row}, and \textit{Move in Order} (\cref{fig:real_world}); the
last two are language-conditioned, with the instruction naming the order in
which the objects are to be handled.
We collect 200 demonstrations per task and jointly train a single policy with a
global batch size of 512, concatenating the three camera views into a single
image as in RoboTwin.
Each task is evaluated over 30 trials on three axes: grasping,
placement, and, for the two tasks that specify one, whether the order
was followed.

\subsection{Main Results}
\label{sec:exp_main}




\begin{wrapfigure}{r}{0.52\textwidth}
  \centering
  \vspace{-\intextsep}
  \includegraphics[width=\linewidth]{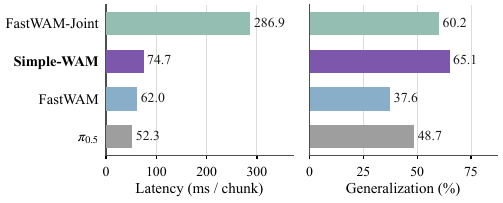}
  \caption{\textbf{Inference Latency.} Per-chunk latency and mean success rate over the four generalization settings of~\cref{tab:main}.}
  \label{fig:efficiency}
\end{wrapfigure}
\noindent\textbf{Simulation Results.}
\cref{tab:main} reports all three axes. Under environmental perturbation
Simple-WAM averages $79.5$ over the seven LIBERO-Plus factors and is best on six among models without embodied pretraining, against $67.7$ for the explicit paradigm and $53.8$ for the latent
one. It comes within $4.9$ points of $\pi_{0.5}$, which is initialized with
large-scale embodied pretraining, and leads every model trained without it by
at least $11.8$.
The margin falls on the four factors that change what the camera sees: a visual
shift corrupts the future the explicit paradigm generates, while a future left
at noise has nothing to corrupt, consistent with Finding~2 on this axis.
With the demonstrations per task reduced it reaches $92.4$ and
$97.2$ on LIBERO at $5$ and $10$ shots and $37.2$ on RoboTwin, comparable to the
explicit paradigm on all three, while the latent one collapses on RoboTwin to
$4.8$. On tasks withheld from training it reaches $10.1$ without any data of
them and $73.6$ with action-free video, against $6.3$ and $69.9$;
the corresponding RoboTwin pairs are $6.2$ against $5.4$ and $44.5$ against
$47.3$; \cref{tab:per_suite} breaks both axes down by suite. The latent paradigm falls behind on every axis, as Finding~1 reports: future modeling is a test-time requirement, not only a training objective.

\begin{wrapfigure}{r}{0.52\textwidth}
  \centering
  \vspace{-\intextsep}
  \includegraphics[width=0.9\linewidth]{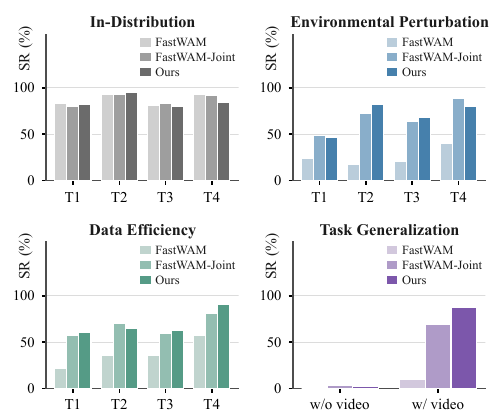}
  \caption{\textbf{Real-world results.} Success rate (\%) over 30 evaluation
  runs. T1--T4 are the four training tasks of~\cref{fig:real_world}, in order;
  task generalization is the held-out \textit{Store in Order}.}
  \label{fig:real_world_results}
\end{wrapfigure}

\noindent\textbf{Efficiency.}
\cref{fig:efficiency} places these gains against their cost. Simple-WAM runs at
$74.7$\,ms per action chunk, $3.8\times$ faster than the explicit paradigm
at $286.9$\,ms and within $12.7$\,ms of the latent one at $62.0$\,ms, while
averaging the best across the four generalization settings.
The additional computation does not yield a consistent improvement.

\begin{figure}[t]
  \centering
  \includegraphics[width=\linewidth]{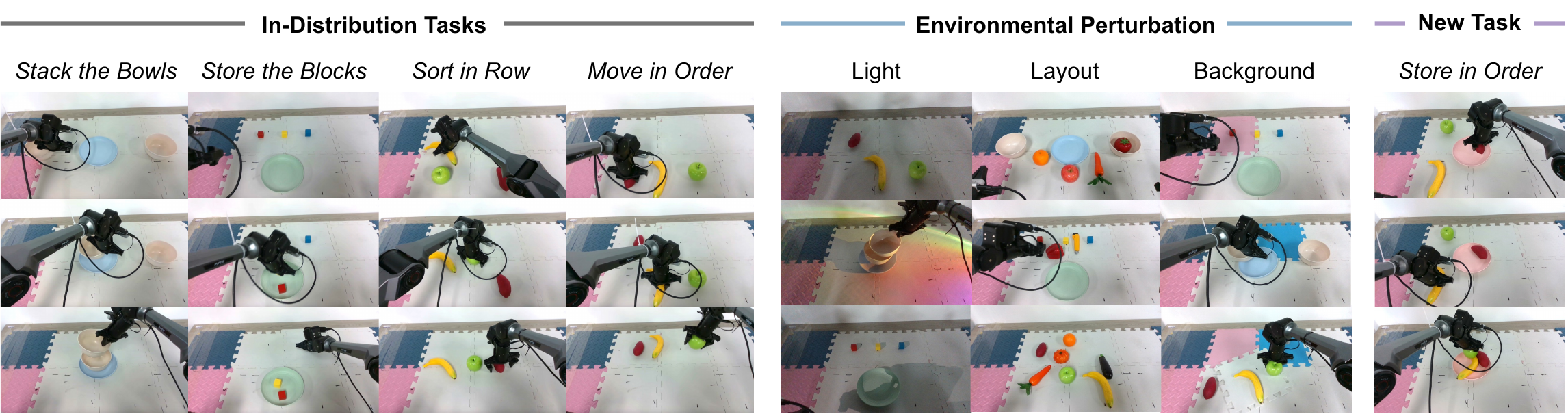}
  \caption{\textbf{The real-world setup.}
  \textbf{Left:} the four training tasks, one per column, each shown at three
  moments of a single rollout.
  \textbf{Middle:} the three environmental perturbations, varying the
  lighting, the object layout and the background.
  \textbf{Right:} \textit{Store in Order}, the held-out task.}
  \label{fig:real_world}
\end{figure}

\noindent\textbf{Real-World Evaluation.}
\cref{fig:real_world_results} reports the four tasks of
\cref{fig:real_world}. In distribution the three models are within
$2.3$ points of each other, reproducing the parity of Finding~1 on real-world tasks. Under the environmental perturbations the latent paradigm falls to $25.2$ while Simple-WAM holds $69.2$, and with the
demonstrations reduced to 50 per task it holds $69.6$ against $37.2$. On held-out
\textit{Store in Order}, action-free video of the task raises Simple-WAM from
$2.2$ to $87.8$, against $68.9$ for the explicit paradigm and $10.0$ for
the latent one. \cref{tab:real_world_full} gives the per-task numbers and their
standard deviations. 

\subsection{Ablation Study}
\label{sec:exp_ablation}

We ablate LIBERO-Spatial along the three generalization axes, using task generalization with-video.

\noindent\textbf{Probability of the mixed schedule.}
\cref{tab:ablation_p} sweeps $p$ over its whole range.
On the three-axis average, $p=0.25$, $0.5$ and $0.75$ differ by $1.3$ points,
showing that performance is robust to the choice of $p$. Both endpoints fall
lower, to $90.8$ and $88.6$.
At $p=0$ training reverts to the explicit paradigm and leaves the mismatch
that~\cref{sec:method_train} mitigates; at $p=1$ the flow time never varies,
and training departs from the objective the video backbone was pretrained
under. Both extremes therefore perform poorly.

\begin{table}[t]
  \centering
  \footnotesize
  \caption{\textbf{Ablations on LIBERO-Spatial.} The
  \colorbox{lightgray}{shaded} entries are the default setting and \textbf{bold} marks the best.}
  \label{tab:ablation}
  \setlength{\tabcolsep}{4.2pt}

  \begin{subtable}[t]{0.48\linewidth}
    \centering
    \caption{Mixture probability $p$}
    \label{tab:ablation_p}
    \begin{tabular}{lcc>{\columncolor{lightgray}}ccc}
      \toprule
      & $0$ & $0.25$ & $0.5$ & $0.75$ & $1$ \\
      \midrule
      Environmental perturbation & 83.3 & \textbf{90.2} & 87.5 & 89.1 & 85.7 \\
      Data efficiency & 97.0 & 97.4 & \textbf{97.8} & 94.4 & 96.8 \\
      Task generalization & 92.2 & 95.8 & \textbf{97.2} & 96.2 & 83.4 \\
      \midrule
      Average & 90.8 & \textbf{94.5} & 94.2 & 93.2 & 88.6 \\
      \bottomrule
    \end{tabular}
  \end{subtable}
  \begin{subtable}[t]{0.48\linewidth}
    \centering
    \caption{Future tokens}
    \label{tab:ablation_token}
    \begin{tabular}{l>{\columncolor{lightgray}}ccc}
      \toprule
      & Noise & Learn. & Zeros \\
      \midrule
      Environmental perturbation & \textbf{87.5} & 59.8 & 65.3 \\
      Data efficiency & \textbf{97.8} & 87.6 & 89.6 \\
      Task generalization & \textbf{97.2} & 10.6 & 9.8 \\
      \midrule
      Average & \textbf{94.2} & 52.7 & 54.9 \\
      \bottomrule
    \end{tabular}
  \end{subtable}
\end{table}

\noindent\textbf{Future video tokens.}
In this part, we investigate whether pure Gaussian noise is a good input for the spatiotemporal information the action expert conditions on.
We replace the future video tokens the action expert reads with learnable query
embeddings or with zeros, and keep the video diffusion target in place so that
the video branch trains as before.
As shown in~\cref{tab:ablation_token}, every axis drops, by $27.7$, $10.2$ and $86.6$ points for the queries and by
$22.2$, $8.2$ and $87.4$ for the zeros. We attribute this to how far the substitution departs from pretraining: the video backbone saw neither in pretraining, while noised video tokens are exactly what it denoises. These results show that our design is simple yet effective because it keeps the original training strategy and thus benefits most from the video pretraining, stressing pretraining-aligned conditioning.

\section{Conclusion}
\label{sec:conclusion}

In this work, we compared the explicit and latent paradigms under a matched protocol. The two match in distribution yet diverge on all three generalization axes, and future video tokens left at pure noise recover almost all of that gap. Simple-WAM therefore conditions on a fully noised future in one pass, leading the explicit paradigm on nearly every measurement at $3.8\times$ its speed. Our conclusions rest on a $5$B backbone without embodied pretraining; whether they hold at larger scale, or with it, remains open.

\bibliography{main}
\bibliographystyle{icml2025}

\newpage
\appendix
\section{Appendix}
\setcounter{table}{0}
\renewcommand{\thetable}{A\arabic{table}}

\subsection{Training Details}
\label{app:training_details}

\cref{tab:hyper} lists our training configuration. The other baselines follow their official settings.

\begin{table}[h]
  \centering
  \footnotesize
  \caption{Training configuration.}
  \label{tab:hyper}
  \setlength{\tabcolsep}{18pt}
  \begin{tabular}{lccc}
    \toprule
    & LIBERO & RoboTwin 2.0 & Real world \\
    \midrule
    Optimizer       & \multicolumn{3}{c}{AdamW, $\beta=(0.9,\,0.95)$} \\
    Learning rate   & \multicolumn{3}{c}{$1\times10^{-4}$} \\
    Weight decay    & \multicolumn{3}{c}{$1\times10^{-2}$} \\
    LR schedule     & \multicolumn{3}{c}{cosine, $5\%$ warmup} \\
    Precision       & \multicolumn{3}{c}{BF16} \\
    \midrule
    Batch size      & $128$          & $1{,}024$      & $512$ \\
    Training steps  & $20$k          & $30$k          & $30$k \\
    Resolution      & $224\times448$ & $384\times320$ & $384\times320$ \\
    \bottomrule
  \end{tabular}
\end{table}

\subsection{Per-Suite Simulation Results}
\label{app:per_suite}

\cref{tab:per_suite} gives the detailed results of~\cref{tab:main} on Data Efficiency and Task Generalization axes. For LIBERO each suite is the held-out fold of the four-fold cross-validation; RoboTwin 2.0 is reported under its clean and randomized scenes, and its data efficiency is measured at 10 demonstrations per task rather than 5.

\begin{table}[h]
  \centering
  \footnotesize
  \caption{Success rate (\%) behind the data-efficiency and
  task-generalization columns of~\cref{tab:main}.}
  \label{tab:per_suite}
  \setlength{\tabcolsep}{7.5pt}
  \begin{tabular}{lccccccccc}
    \toprule
    & \multicolumn{5}{c}{LIBERO} & \multicolumn{3}{c}{RoboTwin 2.0} \\
    \cmidrule(lr){2-6}\cmidrule(lr){7-9}
    & Spatial & Object & Goal & Long & Avg. & Clean & Rand. & Avg. \\
    \midrule
    \multicolumn{9}{l}{\textit{Data efficiency, 5-shot}} \\
    \hspace*{0.6em}\gy{$\pi_{0.5}$} & \gy{92.1} & \gy{94.5} & \gy{89.7} & \gy{73.6} & \gy{87.5} & \gy{--} & \gy{--} & \gy{--} \\
    \hspace*{0.6em}VLA-Adapter & 85.7 & 95.0 & 75.8 & 54.0 & 77.6 & -- & -- & -- \\
    \hspace*{0.6em}\gy{Lingbot-VA} & \gy{59.0} & \gy{96.3} & \gy{88.6} & \gy{70.2} & \gy{78.5} & \gy{--} & \gy{--} & \gy{--} \\
    \hspace*{0.6em}FastWAM-Joint & \textbf{97.0} & \textbf{99.4} & \textbf{92.4} & 77.2 & 91.5 & -- & -- & -- \\
    \hspace*{0.6em}FastWAM & 84.4 & 93.4 & 77.6 & 57.4 & 78.2 & -- & -- & -- \\
    \hspace*{0.6em}Simple-WAM & \textbf{97.0} & \textbf{99.4} & 89.6 & \textbf{83.6} & \textbf{92.4} & -- & -- & -- \\
    \midrule
    \multicolumn{9}{l}{\textit{Data efficiency, 10-shot}} \\
    \hspace*{0.6em}\gy{$\pi_{0.5}$} & \gy{95.4} & \gy{96.6} & \gy{91.0} & \gy{82.8} & \gy{91.5} & \gy{35.5} & \gy{32.3} & \gy{33.9} \\
    \hspace*{0.6em}VLA-Adapter & 92.8 & 92.4 & 89.4 & 69.0 & 85.9 & -- & -- & -- \\
    \hspace*{0.6em}\gy{Lingbot-VA} & \gy{89.1} & \gy{95.8} & \gy{89.2} & \gy{76.4} & \gy{87.6} & \gy{4.1} & \gy{3.6} & \gy{3.9} \\
    \hspace*{0.6em}FastWAM-Joint & \textbf{98.0} & \textbf{98.4} & 97.2 & 94.2 & 97.0 & 36.8 & 33.2 & 35.0 \\
    \hspace*{0.6em}FastWAM & 87.4 & 95.2 & 88.0 & 83.4 & 88.5 & 9.6 & 0.1 & 4.8 \\
    \hspace*{0.6em}Simple-WAM & 97.8 & \textbf{98.4} & \textbf{97.8} & \textbf{94.8} & \textbf{97.2} & \textbf{38.8} & \textbf{35.5} & \textbf{37.2} \\
    \midrule
    \multicolumn{9}{l}{\textit{Task generalization, w/o video}} \\
    \hspace*{0.6em}\gy{$\pi_{0.5}$} & \gy{18.2} & \gy{7.2} & \gy{12.2} & \gy{0.0} & \gy{9.4} & \gy{2.8} & \gy{2.1} & \gy{2.5} \\
    \hspace*{0.6em}VLA-Adapter & 1.2 & 0.0 & 0.0 & 0.0 & 0.3 & -- & -- & -- \\
    \hspace*{0.6em}\gy{Lingbot-VA} & \gy{29.7} & \gy{12.4} & \gy{18.8} & \gy{0.0} & \gy{15.2} & \gy{0.0} & \gy{0.0} & \gy{0.0} \\
    \hspace*{0.6em}FastWAM-Joint & 13.4 & 1.6 & \textbf{10.0} & 0.0 & 6.3 & 6.4 & 4.3 & 5.4 \\
    \hspace*{0.6em}FastWAM & 0.0 & 8.4 & 0.0 & 0.0 & 2.1 & 4.6 & 2.4 & 3.5 \\
    \hspace*{0.6em}Simple-WAM & \textbf{19.8} & \textbf{10.4} & \textbf{10.0} & 0.0 & \textbf{10.1} & \textbf{7.4} & \textbf{4.9} & \textbf{6.2} \\
    \midrule
    \multicolumn{9}{l}{\textit{Task generalization, w/ video}} \\
    \hspace*{0.6em}\gy{$\pi_{0.5}$} & \gy{18.2*} & \gy{7.2*} & \gy{12.2*} & \gy{0.0*} & \gy{9.4*} & \gy{2.8*} & \gy{2.1*} & \gy{2.5*} \\
    \hspace*{0.6em}VLA-Adapter & 1.2* & 0.0* & 0.0* & 0.0* & 0.3* & -- & -- & -- \\
    \hspace*{0.6em}\gy{Lingbot-VA} & \gy{96.2} & \gy{90.3} & \gy{58.2} & \gy{39.6} & \gy{71.1} & \gy{1.2} & \gy{1.0} & \gy{1.1} \\
    \hspace*{0.6em}FastWAM-Joint & 95.4 & 94.8 & \textbf{57.6} & 31.8 & 69.9 & \textbf{47.7} & \textbf{46.8} & \textbf{47.3} \\
    \hspace*{0.6em}FastWAM & 6.2 & 17.4 & 0.0 & 0.0 & 5.9 & 4.3 & 5.3 & 4.8 \\
    \hspace*{0.6em}Simple-WAM & \textbf{97.2} & \textbf{99.2} & 56.4 & \textbf{41.4} & \textbf{73.6} & 45.1 & 43.8 & 44.5 \\
    \bottomrule
  \end{tabular}
\end{table}

\subsection{Per-Task Real-World Results}
\label{app:real_world}

\cref{tab:real_world_full} reports the mean and standard deviation of real-world results, where T1--T4 are the four training tasks of~\cref{fig:real_world} and the held-out task is evaluated without and with action-free video of it. A run is not a binary success: it receives the fraction of the sub-goals listed in~\cref{sec:exp_setup} that the policy completes, and each cell averages thirty such runs. ``Avg.'' averages the four task means and is the quantity quoted in~\cref{sec:exp_main}.

\begin{table}[h]
  \centering
  \footnotesize
  \caption{Real-world results per task.}
  \label{tab:real_world_full}
  \setlength{\tabcolsep}{10pt}
  \begin{tabular}{lccccc}
    \toprule
    & T1 & T2 & T3 & T4 & Avg. \\
    \midrule
    \multicolumn{6}{l}{\textit{In-distribution}} \\
    \hspace*{0.6em}FastWAM & $83.3${\scriptsize$\pm$19.2} & $92.5${\scriptsize$\pm$16.9} & $81.1${\scriptsize$\pm$12.9} & $93.3${\scriptsize$\pm$14.1} & $87.6$ \\
    \hspace*{0.6em}FastWAM-Joint & $80.0${\scriptsize$\pm$21.9} & $92.5${\scriptsize$\pm$16.9} & $83.3${\scriptsize$\pm$13.1} & $92.2${\scriptsize$\pm$11.8} & $87.0$ \\
    \hspace*{0.6em}Simple-WAM & $81.7${\scriptsize$\pm$16.6} & $95.0${\scriptsize$\pm$10.5} & $80.0${\scriptsize$\pm$15.5} & $84.4${\scriptsize$\pm$18.3} & $85.3$ \\
    \midrule
    \multicolumn{6}{l}{\textit{Environmental perturbation}} \\
    \hspace*{0.6em}FastWAM & $23.3${\scriptsize$\pm$26.3} & $17.5${\scriptsize$\pm$23.7} & $20.0${\scriptsize$\pm$13.7} & $40.0${\scriptsize$\pm$23.5} & $25.2$ \\
    \hspace*{0.6em}FastWAM-Joint & $48.3${\scriptsize$\pm$16.6} & $72.5${\scriptsize$\pm$29.9} & $63.3${\scriptsize$\pm$35.2} & $88.9${\scriptsize$\pm$21.0} & $68.3$ \\
    \hspace*{0.6em}Simple-WAM & $46.7${\scriptsize$\pm$18.9} & $82.5${\scriptsize$\pm$16.9} & $67.8${\scriptsize$\pm$22.5} & $80.0${\scriptsize$\pm$16.4} & $69.2$ \\
    \midrule
    \multicolumn{6}{l}{\textit{Data efficiency}} \\
    \hspace*{0.6em}FastWAM & $21.7${\scriptsize$\pm$20.9} & $35.0${\scriptsize$\pm$26.9} & $35.6${\scriptsize$\pm$18.0} & $56.7${\scriptsize$\pm$32.5} & $37.2$ \\
    \hspace*{0.6em}FastWAM-Joint & $56.7${\scriptsize$\pm$22.5} & $70.0${\scriptsize$\pm$30.7} & $58.9${\scriptsize$\pm$19.6} & $81.1${\scriptsize$\pm$18.2} & $66.7$ \\
    \hspace*{0.6em}Simple-WAM & $60.0${\scriptsize$\pm$21.1} & $65.0${\scriptsize$\pm$33.7} & $62.2${\scriptsize$\pm$32.4} & $91.1${\scriptsize$\pm$11.5} & $69.6$ \\
    \midrule
    \multicolumn{6}{l}{\textit{Task generalization, held-out task}} \\
    & w/o video & w/ video & & & \\
    \hspace*{0.6em}FastWAM & $0.0${\scriptsize$\pm$0.0} & $10.0${\scriptsize$\pm$9.7} & & & \\
    \hspace*{0.6em}FastWAM-Joint & $3.3${\scriptsize$\pm$5.4} & $68.9${\scriptsize$\pm$20.2} & & & \\
    \hspace*{0.6em}Simple-WAM & $2.2${\scriptsize$\pm$4.7} & $87.8${\scriptsize$\pm$19.2} & & & \\
    \bottomrule
  \end{tabular}
\end{table}

\end{document}